\documentclass[journal]{IEEEtran}

\usepackage{silence}
\usepackage{array}
\usepackage[caption=false,font=normalsize,labelfont=sf,textfont=sf]{subfig}
\usepackage{textcomp}
\usepackage{stfloats}
\usepackage{verbatim}
\def\BibTeX{{\rm B\kern-.05em{\sc i\kern-.025em b}\kern-.08em
    T\kern-.1667em\lower.7ex\hbox{E}\kern-.125emX}}
\usepackage{balance}

\usepackage[numbers]{natbib}
\usepackage{multicol}
\usepackage[bookmarks=true]{hyperref}
\usepackage[utf8]{inputenc} 
\usepackage[T1]{fontenc}    
\usepackage{hyperref}       
\usepackage{booktabs}       
\usepackage{nicefrac}       
\usepackage{microtype}      
\usepackage{lipsum}
\usepackage{graphicx}
\graphicspath{ {./images/} }
\usepackage{hyperref}
\hypersetup{
    colorlinks=true,  
    urlcolor=cyan,
    citecolor=black,
}

\usepackage{times}  
\usepackage{helvet}  
\usepackage{threeparttable}
\usepackage{courier}  
\usepackage{graphicx} 
\usepackage{caption} 
\usepackage[dvipsnames,svgnames,table]{xcolor}

\usepackage{algorithm}
\usepackage{algpseudocode}
\usepackage{booktabs}
\usepackage{flushend}
\usepackage{amssymb}
\usepackage{xurl}
\usepackage{pifont}
\usepackage{amsmath}
\usepackage{amsfonts,bm} 
\usepackage{tikz}
\newcommand{\cmark}{\ding{51}}%
\newcommand{\xmark}{\ding{55}}%
\usepackage{amsthm}
\usepackage{soul}
\usepackage[utf8]{inputenc}

\newcommand{\eg}{e.\,g.}

\begin{document}
\title{Affordance-based Robot Manipulation with Flow Matching}
\author{Fan Zhang and Michael Gienger
\\ Honda Research Institute EU
\\ Email: firstname.lastname@honda-ri.de
}

\maketitle

\begin{abstract}
We present a framework for assistive robot manipulation that addresses two fundamental challenges: efficient adaptation of large-scale models for scene affordance understanding and effective learning of robot actions by grounding the visual affordance. To tackle the first challenge, we adopt a parameter-efficient prompt tuning method, prepending learnable text prompts to a frozen vision model to predict affordances, while considering spatial and semantic relationships in multi-task scenarios. For the second challenge, we propose a flow matching method, representing a robot visuomotor policy as a conditional process of flowing random waypoints to desired robot actions. We introduce a real-world dataset with 10 tasks to evaluate our approach. Experiments show our prompt tuning method achieves competitive or superior performance to other finetuning protocols across data scales, while satisfying parameter efficiency.  Flow matching yields more stable training and faster inference, while maintaining comparable generalization performance to diffusion policy. Our framework seamlessly unifies parameter-efficient affordance learning and robot action generation with flow matching. 
\url{https://hri-eu.github.io/flow-matching-policy/}
\end{abstract}

\begin{IEEEkeywords}
Visual Learning, Flow Matching, Affordance Learning, Human-Centered Robotics
\end{IEEEkeywords}

\section{Introduction}
\label{sec:Introduction}
\IEEEPARstart{R}{ecent} advances in vision-language models (VLMs) present unprecedented opportunities to solve robot manipulation problems. Attempts in the field have focused on three primary aspects: 1) End-to-end learning manipulation from scratch. These approaches~\citep{padalkar2023open} make the fewest assumptions on tasks and are formulated in language-image-to-action prediction models. 2) Off-the-shelf-vision-language models for robot manipulation. These works have prompted pre-trained VLMs in various contexts of robot motion learning, including reward design for reinforcement learning~\citep{ma2023eureka}, python coding~\citep{liang2023code}, joint actions~\citep{wang2023prompt}, etc. 3) Intermediate substrate to bridge high-level language-image instructions and low-level robot policies. These works usually introduce some form of prior derived from human knowledge as an intermediate stage to alleviate the sample inefficiency problem of end-to-end learning, including affordances~\citep{huang2023voxposer}, primitive skills~\citep{ingelhag2024robotic}, etc. In this paper, we follow the third line of work to unify a parameter-efficient affordance model and a low-level robot flow matching policy.

\begin{figure*}[!t]
\centering
\includegraphics[width=\textwidth]{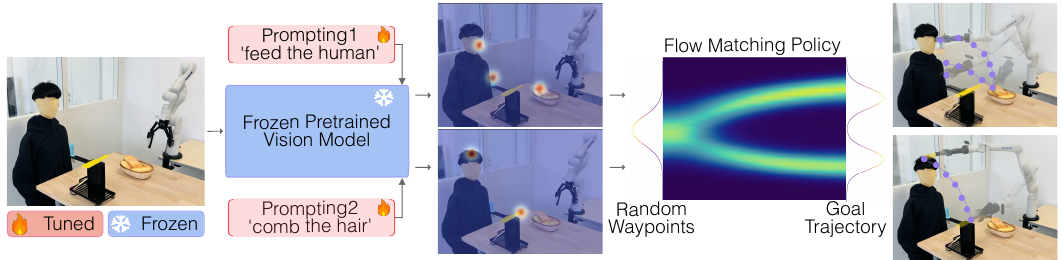}
\caption{The proposed framework of unifying affordance map learning and action generation for robot manipulation. Given the same visual scene with different language instructions, the model first extracts instruction-relevant manipulation affordances. This is achieved through a prompt tuning method that prepends learnable text-conditioned prompts in a frozen vision foundation model. Then, a flow matching policy is proposed to transform the random waypoints to the desired action trajectories, guided by task-relevant affordance maps.}
\label{fig:overall}
\end{figure*}

Extracting affordance knowledge has long inspired the robot community~\citep{xiao2022masked}. Recent state-of-the-art works have proposed the affordance-based robot policy by describing affordances with 6D poses or 2D point tracks~\citep {nasiriany2024rt, bharadhwaj2024track2act}. For example, ~\citep {nasiriany2024rt} represents affordances as the relative poses between the robot end effector and the object at key stages of the task. ~\citep {bharadhwaj2024track2act} uses 2D keypoint track predictions to infer a sequence of rigid transforms of the object affordance to be manipulated. Other research decouples affordance learning and robot policy by using more straightforward affordance representations like 2D masks or keypoints~\citep{stone2023open, yuan2024robopoint}. These works typically localize task-specific affordances directly on objects relevant to certain actions, given detailed text instructions. Our affordance model additionally considers the relational context needed for interactions, including the spatial and semantic functional interplay relationship between multiple elements. We concentrate on multi-task scenarios with text prompting. As shown in Fig.~\ref{fig:overall}, given the same visual scene but with different language instructions, we aim to extract different affordances for robot policy learning through our proposed model. For example, given a simple instruction of `feeding the human', a sequence of 2D affordance heatmaps on a spoon handle, food and a person's mouth are identified. 

Recent affordance models have successfully fine-tuned pre-trained vision-language foundation models to extract affordance knowledge~\citep{qian2024affordancellm, yuan2024robopoint}. To leverage the capabilities of pre-trained foundation models while simultaneously mitigating the associated computational costs, some works have explored parameter-efficient fine-tuning (PEFT) large vision-language models~\citep{jia2022visual}. One representative line of PEFT research has concentrated on prompt tuning methods, which prepend learnable prompts to a large frozen pre-trained model and optimize them via gradients during finetuning. In contrast to studies in NLP, it has been shown that prompt tuning could match or even outperform full fine-tuning and adapter-based methods, but with substantially less parameter storage in visual domains~\citep{zhu2023visual}. Inspired in part by the notion of human cognitive penetrability mechanism~\citep{maier2019no} that uses linguistic knowledge to tune ongoing visual processing, we aim to incorporate learnable text-conditioned prompts into any vision foundation model while keeping it frozen, preserving its visual understanding capabilities, to learn instruction-relevant manipulation affordance maps.

The subsequent challenge involves deploying the visual affordance across robot manipulation learning paradigms. A recent line of work builds on successes in curriculum-based imitation learning~\citep{blessing2023information}, probabilistic movement primitive modeling~\citep{yildirim2024conditional}, and diffusion models~\citep{chi2023diffusion} to generate motion trajectories to capture multimodal action distributions. Flow Matching is another novel generative method. Sharing theoretical similarities with diffusion process, flow matching aims to regress onto a deterministic vector field to flow samples toward the target distribution. \citep{lipman2022flow} has proven that the simplicity of flow matching objectives allows favorable performance in stable training and generation quality compared to solving complex stochastic differential equations in the stochastic denoising diffusion process. We extend flow matching to the robotics domain. As shown in Fig.~\ref{fig:overall}, the proposed method would flow the random waypoints to the desired action trajectories based on multi-task affordances in a single flow matching policy.

We also construct a Real-world Activities of Daily Living (ADLs) dataset with 10 tasks. The novelty of our dataset is that it contains the same scenarios but with multi-task affordance and robot trajectories. Experimental evaluation on our dataset empirically demonstrates that the prompt tuning method for learning affordances achieves performance competitive, and sometimes beyond other finetuning protocols across data scales and vision-language fusion architectures. 

This work seamlessly grounds VLM-based affordance with flow matching for real-world robot manipulation. We leverage the capability of the flow matching policy to represent multi-modal action distributions for learning accurate 6D robot actions, while 2D affordance maps readily provide sufficient guidance in shaping the policy. Depending on the system design, our approach can run at approximately 30 Hz (affordance learning: 18.340 ms; flow matching with one-step inference: 13.228 ms), making it suitable for closed-loop manipulation tasks. We have further systematically evaluated robot manipulation with flow matching on several benchmarks, including various input representations, robot control types, and manipulation tasks. We showcase that across several benchmarks, flow matching attains favorable performance in training stability, generation quality, and computational efficiency amongst competing methods of behavior cloning.  Our code is publicly available.

The main contributions can be summarized as follows:
\begin{enumerate}
\item A parameter-efficient prompt tuning method for adapting pretrained vision foundation model conditioned on language instructions to learn manipulation affordances, incorporating the spatial and semantic interactions between elements in multi-task scenarios.
\item A novel formulation using flow matching for closed-loop 6D robot manipulation learning from various inputs, including visual affordances. Empirical and extensive results show that flow matching leads to consistently favorable results than some alternative behavior cloning methods in various manipulation tasks. This includes \textbf{more stable training and evaluation and noticeably faster inference, while maintaining comparable generalization performance to diffusion policy (DDIM, DDPM and score-based models)}. 
\item A framework seamlessly unifies parameter-efficient affordance learning and robot action generation with flow matching. Experimental results, compared against end-to-end learning and off-the-shelf VLM-based trajectory generation frameworks, demonstrate that affordance representations provide consistent guidance for shaping manipulation policies.
\end{enumerate}

\section{Related Work}
\label{sec:Related}

\subsection{Affordance Learning and Affordance-Based Robot Policy}
Humans rely heavily on affordances to perform day-to-day tasks across environments efficiently. The concept of affordance has been introduced in~\citep{gibson2014ecological}, referring to the ability to perform certain actions with objects in the context of a given scene. Several state-of-the-art works have adopted this concept and successfully proposed affordance-based robot policy by learning the affordance parameterization that could be amenable to deployment on robots, including 6D pose~\citep{nasiriany2024rt}, stable object placement~\citep{zeng2021transporter}, point tracks~\citep{bharadhwaj2024track2act}, and post-contact trajectory~\citep{bahl2023affordances}. Other approaches decouple actionable representation for affordances from robot policy by using more straightforward affordance representations, like 2D masks~\citep{stone2023open}, keypoints~\citep{manuelli2019kpam}, heatmaps~\citep{zhang2022learning}, part segmentation~\citep{do2018affordancenet}, dense image feature descriptors~\citep{florence2018dense}. We follow this idea of decoupling and use 2D heatmaps as affordances. 

Recent SoA works have successfully leveraged pre-trained vision-language models (\eg, LLaVA and Vicuna) to extract affordance knowledge, as in AffordanceLLM~\citep{qian2024affordancellm} and Robopoint~\citep{yuan2024robopoint}. This raises three important research questions: i) How can language-vision inputs for instruction-aware vision encoding be readily fused? ii) How to parameter-efficiently finetune a pretrained foundation model? and iii) How to encode the spatial and semantic functional interplay relationship between elements in multi-task scenarios for robot manipulation affordance learning? We introduce a prompt tuning method as a solution to these questions.

\subsection{Parameter-Efficient Finetuning}
Given the dominance of large-scale vision-language models, many approaches have been proposed to efficiently finetune a frozen pretrained model for downstream tasks to speed up training and reduce memory~\citep{gupta2022towards, radford2021learning}. Two representative parameter-efficient finetuning methods are adapters and prompting. The adapter-based research designs the adapter variants that could add extra lightweight modules~\citep{gao2024clip, hu2021lora, liu2023tail, sharma2023lossless}. Other work focuses on prompt tuning~\citep{liu2021p, liu2023pre}, which treats learnable prompts as continuous vectors and computes their gradients with backpropagation during training. Studies on randomly generalised trainable prompts~\citep{jia2022visual} for universal use or condition-admitted prompt variables~\citep{sohn2023visual} for better specific task performance have both been explored. The extension of prompt tuning to vision tasks has gained massive success. Visual prompt tuning~\citep{jia2022visual} has manipulated visual prompts to steer models in arbitrary vision tasks. Dancemvp~\citep{zhong2024dancemvp, zhong2023contrastive}  has adjusted language prompts to guide text-audio models in performing dancing assessment tasks. Inspired by its recent success, we extend the prompt tuning technologies to address the challenge of adapting large pretrained vision-language models to affordance learning for robot manipulation. The intuition is clear: if the model understands the posed text instruction and the inherent context, it should extract visual affordances that directly correspond to the relevant image aspects. Our method achieves the above goal by integrating learnable text-conditioned prompts into a large vision encoder, while keeping it frozen to preserve visual understanding capabilities. Besides, recent research on VLM for affordance-based robot policy typically localizes task-specific affordances directly on objects relevant to certain actions, which requires detailed language instructions about all task-related objects and actions. We aim to enable our affordance model to incorporate relational context essential for interactions, capturing both spatial and semantic-functional relationships among multiple objects and persons, where only simple prompts are required.

\subsection{Robot Learning from Demonstration}
Imitation learning has been a common paradigm for robots, which requires simulated or real-world demonstration data collection~\citep{levine2016end}. To improve data efficiency, extensive work has been proposed to learn robot policies on top of visual representations~\citep{liu2024moka}, such as keypoints or affordance heatmaps~\citep{liu2024moka}, instead of end-to-end raw images~\citep{goyal2023rvt}. This paper concentrates on using affordances to guide low-level robot manipulation. In terms of network architectures for robot learning, prior works have successfully investigated convolutional networks~\citep{levine2016end}, Transformers~\citep{sridhar2023memory}, generative adversarial networks~\citep{ho2016generative}, Energy-Based Models~\citep{florence2022implicit}, etc. However, the collected data is usually expected to be non-convex and multi-modal due to the variability in human demonstrations. Recent works have addressed this problem by reformulating the robot policy as a generative process. Diffusion policy~\citep{chi2023diffusion}, including Denoising Diffusion Probabilistic Models (DDPM~\citep{ho2020denoising}), Denoising Diffusion Implicit Models (DDIM~\citep{song2020denoising}) and score-based models~\citep{reuss2023goal},  has emerged as a powerful class of generative models for behavior cloning by representing a robot’s visuomotor policy as a conditional denoising diffusion process. In this work, we investigate flow matching~\citep{lipman2022flow}, a novel generative model that has demonstrated its superiority in image generation, but is much less explored in robotics domains.

\subsection{Flow Matching in Robotics}
Despite its recent progress in image generation~\citep{albergo2022building}, the application of flow matching in robotics domains remains underexplored~\citep{hu2024adaflow, rouxel2024flow}. Few prior studies have concentrated exclusively on certain robot scenarios for deploying flow matching, for pointcloud environment~\citep{chisari2024learning}, dynamics~\citep{nguyen2025flowmp}, Riemannian manifolds~\citep{braun2024riemannian}, etc. We propose to use flow matching to learn multi-task robot behaviors from raw observations, including visual affordances obtained from a vision-language model, in a single supervised policy. We have first systematically evaluated robot manipulation with flow matching on several benchmarks, including various input representations, robot control types, and manipulation tasks.

\section{Methods}

\subsection{Prompt Tuning for Affordance Map Learning}
\begin{figure}[!t]
\centering
\includegraphics[width=0.98\columnwidth]{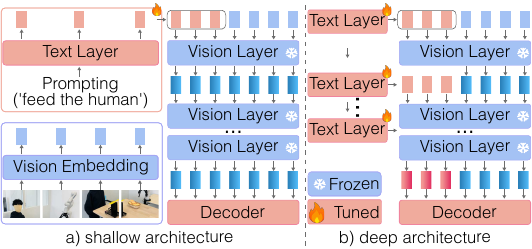}
\caption{Overview of prompt tuning structures used for affordance learning. (Left) For the shallow structure, text-conditioned prompts are prepended to the first vision transformer layer. (Right) For the deep structure, prompts are inserted into every vision layer. Only the prompt-related layers and the decoder are being updated during the training, while the vision transformer remains frozen.}
\label{fig:vpt}
\end{figure}

Providing any type of pre-trained vision transformer, our objective is to learn a set of text-conditioned prompts to maximize the likelihood of correct affordance labels, as shown in Fig.~\ref{fig:vpt}. Only the prompt-related layers and the decoder are being updated during the training, while the vision transformer remains frozen. Inspired by Vision Prompt Tuning~\citep{jia2022visual}, we propose two frameworks: shallow and deep network architectures.

\subsubsection{Shallow Architecture}
The vision transformer layer takes the image patch embeddings $\bm {E}_0$ as input and passes through various layers $\bm {L}_i^v$ to achieve vision features $\bm {E}_i$, where $\bm {E}_i \in \mathbb{R}^{M \times C}$ and $C$ is the channel dimension.
\begin{equation}
\bm {E}_i = \bm {L}_i^v(\bm {E}_{i-1}) \qquad i=1,2,\cdots, N
\nonumber
\end{equation}
Similarly, the text transformer layer could be represented as 
\begin{equation}
\bm {P}_i = \bm {L}_i^p(\bm {P}_{i-1}) \qquad i=1,2,\cdots, N
\nonumber
\end{equation}
where $\bm {P}_0$ denotes the text tokens, text features $\bm {P}_i$ are obtained through various layers $\bm {L}_i^P$, where $\bm {p}_i \in \mathbb{R}^{K \times C}$.

As shown in Fig.~\ref{fig:vpt}, for the shallow structure, only one text transformer layer is used to compute text features $\bm {P}_1$, which are then treated as prompts and inserted into the first vision transformer Layer:
\begin{align*}\notag
[\bm{Z}_1, \bm{E}_1] &= \bm {L}_1^v([\bm {P}_{1}, \bm {E}_{0}]) \\
[\bm{Z}_i, \bm{E}_i] &= \bm {L}_i^v([\bm {Z}_{i-1}, \bm {E}_{i-1}]) 
\end{align*}
Then a decoder is added on the global output flattened token sequence to generate visual affordance tokens. 
\begin{equation}
\text{Affordance} = \text{Decoder}(\bm{Z}_N, \bm{E}_N)
\nonumber
\end{equation}

\subsubsection{Deep Architecture} 
For the deep architecture, the only difference is that text features $\bm {P}_i$ are computed through each layer and introduced at the corresponding vision transformer layer's input space:
\begin{align*}\notag
[\_, \bm{E}_1] &= \bm {L}_1^v([\bm {P}_{1}, \bm {E}_{0}]) \\
[\_, \bm{E}_i] &= \bm {L}_i^v([\bm {P}_{i}, \bm {E}_{i-1}]) 
\end{align*}

This is different to parallel adapters, which introduce additional trainable modules that run in parallel with the main transformer layers instead of modifying existing layers.

\begin{figure*}[!t]
\centering
\includegraphics[width=0.98\textwidth]{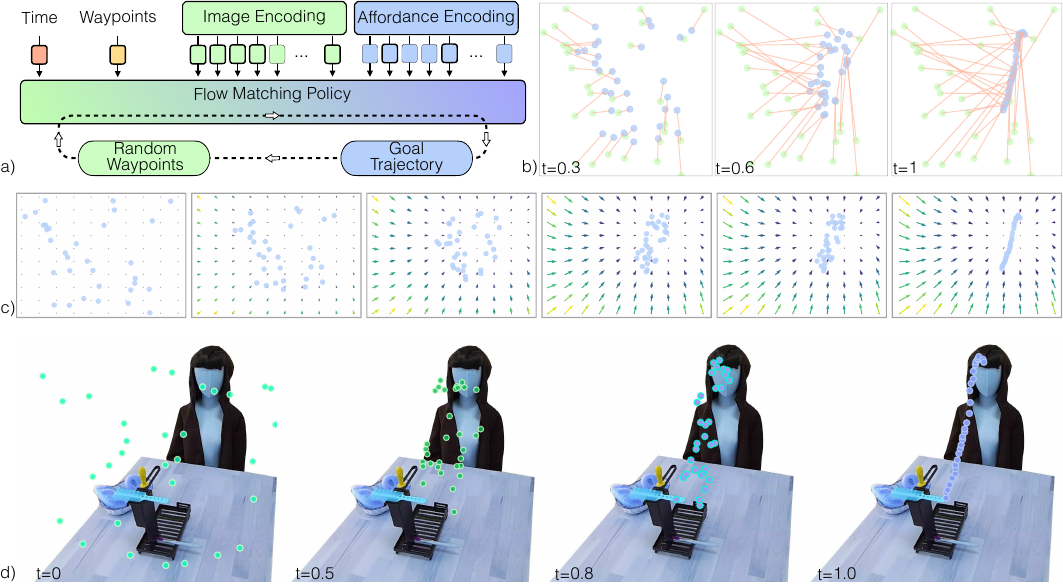}
\caption{Framework of flow matching policy. (a) General formulation. At each time step, flow matching takes visual observation  $\bm{o}$ (\eg, state-based inputs, RGB-D images, visual affordances) as input, and outputs robot actions (\eg, 6D robot end-effector trajectories, robot joint actions, gripper actions). (b-d) Visualization of the inference process of transforming random waypoints to target actions over time from 0 (green) to 1 (purple). Red lines in (b) denote the flow paths. Vector fields are shown in (c).}
\label{fig:flow}
\end{figure*}

\subsubsection{Implementation Details}
In our implementation, we select and adopt a single structure from the two designs. The deep structure demonstrates better performance than the shallow structure, albeit with increased computational cost. Our goal is to integrate textual representations into any vision encoder while keeping it frozen, preserving its visual understanding capabilities. Thus, we have chosen the most basic vision backbone, a pretrained ViT-B-16 transformer. We use the classic CLIP Transformer layers~\citep{radford2021learning} that output 76 tokens. As suggested by MAE~\citep{he2022masked}, the decoder is only used for downstream tasks and could be flexible and lightweight. Thus we use one single transformer decoder layer. 

We use the L2 Mean Squared loss between the predicted and ground truth affordances for network training. The training parameters for the prompt tuning network include an image size of 224 × 224, AdamW with a learning rate of 1.5e-4, including Warmup with step-decay, and a batch size of 256. We add positional embeddings to all the image and language tokens to preserve the positional information. In the subsequent experiments, we will further ablate multiple model variants, including text and vision fusion structures, prompt depth, pretrained weights for vision transformers, etc.

\subsection{Flow Matching Policy}
We build the robot behavioral cloning policy as a generative process of flow matching, which constructs a flow vector that continuously transforms a source probability distribution toward a destination distribution. Flow matching leverages an ordinary differential equation to deterministically mold data distribution, contrasting with Denoising Diffusion Probabilistic Models (DDPM), which is based on a stochastic differential equation by introducing noise. 

\subsubsection{Flow Matching Model}
Given a conditional probability density path $p_t(\bm x | \bm z)$ and a corresponding conditional vector field $\bm u_t(\bm x | \bm z)$, the objective loss of flow matching could be described as:
\begin{equation}
\mathcal{L}_{\text{FM}}(\bm \theta) = \mathbb{E}_{t, q(\bm z), p_t(\bm x | \bm z)} \left\| \mathbf{v}_t(\bm x, \bm \theta) - \mathbf{u}_t(\bm x | \bm z) \right\|^2
\label{eq:loss}
\end{equation}
where $\bm x \sim p_t(\bm x | \bm z)$, $t \sim \mathcal{U}[0, 1]$. Flow matching aims to regress $\mathbf{u}_t(\bm x | \bm z)$ with a time-dependent vector field of flow $\mathbf{v}_t(\bm x, \bm \theta)$ parameterized as a neural network with weights $\bm \theta$. $\mathbf{u}_t(\bm x | \bm z)$ can be further simplified as:
\begin{equation}
\mathbf{u}_t(\bm x | \bm z) = \bm x_1 - \bm x_0 \qquad \bm x_0 \sim p_0, \bm x_1 \sim p_1
\nonumber
\end{equation}
$p_0$ represents a simple base density at time $t=0$, $p_1$ denotes the target complicated distribution at time $t=1$, $\bm x_0$ and $\bm x_1$ are the corresponding samplings. $\mathbf{v}_t(\bm x, \bm \theta)$ is described as: 
\begin{equation}
\mathbf{v}_t(\bm x, \bm \theta) = v_{\bm \theta}(\bm x_t, t)
\label{eq:flow}
\end{equation}
where we define $\bm x_t$ as the linear interpolation between $\bm x_0$ and $\bm x_1$ with respect to time $\bm x_t = t\bm x_1+(1-t)\bm x_0$, following the linear conditional flow theory~\citep{peyre2019computational}. $v_{\bm \theta}$ is a network of the flow model. Thus Equation~(\ref{eq:loss}) could be reformatted as 
\begin{equation}
\mathcal{L}_{\text{FM}}(\bm \theta) = \mathbb{E}_{t, \sim p_0, \sim p_1} \left\| v_{\bm \theta}(\bm x_t, t) -(\bm x_1 - \bm x_0)) \right\|^2
\label{eq:newloss}
\end{equation}
This represents the progression of the scalar flow that transforms data from source to target between time 0 and 1.

\subsubsection{Flow Matching for Visuomotor Policy Learning}
We extend flow matching to learn robot visuomotor policies. This requires two modifications in the formulation: i) modeling the flow estimation conditioned on input observations $\bm o$; ii) changing the output $\bm x$ to represent robot actions. Fig.~\ref{fig:flow} illustrates our model structures.

\textbf{Visual observation Conditioning:} We modify Equation~(\ref{eq:flow}) to allow the model to predict actions conditioned on observations:
\begin{equation}
\mathbf{v}_t(\bm x | \bm{o}) = v_{\bm \theta}(\bm x_t, t| \bm{o})
\nonumber
\end{equation}

\renewcommand{\algorithmicrequire}{\textbf{Input:}}
\renewcommand{\algorithmicensure}{\textbf{Output:}}
\begin{algorithm}[t]
\caption{Robot Flow Matching Policy Training}\label{alg:flow}
\begin{algorithmic}[1]
\Require observation $\bm o$, target robot actions $\bm x_{1}$, source random waypoints $p_0$
\Ensure flow $\mathbf{v}_{\bm \theta}$
\While{not converged}
\State $\bm x_0 \sim p_0$, sample random robot waypoints
\State $t \sim \mathcal{U}[0, 1]$, sample time steps
\State $\bm x_t = t\bm x_1+(1-t)\bm x_0$, linear interpolation
\State $\mathbf{v}_t(\bm x | \bm{o}) = v_{\bm \theta}(\bm x_t, t| \bm{o})$, flow estimation
\State $\nabla_{\bm \theta} \left\| v_{\bm \theta}(\bm x_t, t | \bm{o}) - \dot{\bm{x}}_{t} \right\|$, gradient step
\EndWhile
\State Stopping criteria: training epochs reached
\end{algorithmic}
\end{algorithm}

\textbf{Closed-loop action trajectory prediction:} We execute the action trajectory prediction obtained by our flow matching model for a fixed duration before replanning. At each step, the policy takes the observation data $\bm o$ as input and predicts $Tp$ steps of actions, of which $Ta$ steps of actions are executed on the robot without re-planning. $Tp$ is the action prediction horizon and $Ta$ is the action execution horizon. The whole training process of flow matching is illustrated in Algorithm~\ref{alg:flow}. 

\textbf{Inference:} For the inference procedure, random waypoints are sampled from the source distribution and then flowed into the target trajectory by estimating the flow from $t = 0$ to $t = 1$ over steps. We could use multiple steps $1/\Delta t$ for inference:
\begin{equation}
\bm x_{t+ \Delta t} = \bm x_{t}+\Delta t f(\bm x_t, t| \bm{o}),  \qquad \text{for} \  t \in [0, 1]
\label{eq:inference}
\end{equation}

\begin{algorithm}[t]
\caption{Robot Diffusion Policy (DDPM) Training}\label{alg:ddpm}
\begin{algorithmic}[1]
\Require observation $\bm o$, target robot actions $\bm x_{1}$, source Gaussian noises $p_0$
\Ensure noise $\mathbf{\epsilon}_{\bm \theta}$
\While{not converged}
\State $\bm x_0 \sim p_0$, sample Gaussian noises
\State $t \sim \mathcal{U}[0, 1]$, sample time steps
\State $\bm x_t = \mathcal{N}(\mathbf{x}_t; \sqrt{\bar{\alpha}_t} \mathbf{x}_0, (1 - \bar{\alpha}_t)\mathbf{I})$, forward process
\State $\mathbf{\epsilon}_t(\bm x | \bm{o}) = \epsilon_{\bm \theta}(\bm x_t, t| \bm{o})$, noise estimation
\State $\nabla_{\bm \theta} \left\| \epsilon_{\bm \theta}(\bm x_t, t | \bm{o}) - \bm{\epsilon}_{t} \right\|$, gradient step
\EndWhile
\State Stopping criteria: training epochs reached
\end{algorithmic}
\end{algorithm}

\subsubsection{Implementation Details}
For the network structures of flow matching, we first use ResNet~\citep{he2016deep} for visual embeddings $\bm{o}$. The flow model $f_{\bm \theta}$ is represented with U-Net~\citep{ronneberger2015u}. The flow model predicts vectors $\mathbf{v}_t$ conditioned on visual observation embeddings $\bm{o}$ with Feature-wise Linear Modulation (FiLM)~\citep{perez2018film} as well as the interpolated waypoints $\bm x_t$. In the subsequent experiments, we will further study multiple model variants, including a transformer-based structure, trajectory representation.

In our case of robot manipulation, $\bm x_1$ in Equation~(\ref{eq:newloss}) represents the demonstration robot action trajectories. $\bm x_0$ is the random generated waypoints following a multivariate normal distribution $\bm x_0 \sim \mathcal{N}(0, I)$. $\bm x$ here could denote 6D robot end-effector trajectories, robot joint actions, gripper actions, etc. The visual embeddings $\bm{o}$ include various types of inputs, such as state-based inputs, RGB-D images, and visual affordances.

\subsubsection{Comparisons against Diffusion Policy}
In this section, we provide some insights and intuitions about flow matching and its comparisons against diffusion policy for clarification. Algorithm~\ref{alg:flow} and Algorithm~\ref{alg:ddpm} respectively shows the pseudocode of training flow matching and diffusion policy with DDPM. We can see several similarities and differences between these two methods:
\begin{itemize}
[
    \setlength{\IEEElabelindent}{\dimexpr-\labelwidth-\labelsep}
    \setlength{\itemindent}{\dimexpr\labelwidth+\labelsep}
    \setlength{\listparindent}{\parindent}
]
\item \textbf{Sampling}: By solving a Stochastic Differential Equation (SDE), DDPM generates a clean sample from Gaussian noise. Flow matching regresses onto a target flow vector field that generates a deterministic mapping from source to target data distributions by solving an ordinary differential equation (ODE), which drops all the Gaussian assumptions.
\item \textbf{Probability path}: Flow matching includes a particularly interesting family of probability paths: the conditional vector field with linear interpolant. Flow matching paths with linear interpolation are simpler than diffusion paths, forming straighter trajectories, whereas diffusion paths result in curved paths. These properties seem to empirically translate to more stable training, faster generation, and better performance.
\item \textbf{Reparameterization}: ~\citep{lu2024simplifying, nakkiran2024step} claims that diffusion models and flow matching can be considered equivalent with reparameterization, including rescaling time and reparameterizing the marginal velocity field via the score function. Specifically, ~\citep{nakkiran2024step} argues that DDIM is equivalent to flow-matching with the following parameters:
\begin{align*}
\bm{v}_t^{[\bm{x}_1, \bm{x}_0]}(\bm{x}_t) &:= \frac{1}{2t}(\bm{x}_t - \bm{x}_0) & \text{DDIM} \\
\bm{v}_t^{[\bm{x}_1, \bm{x}_0]}(\bm{x}_t) &:= \frac{1}{2\sqrt{t}}(\bm{x}_0 - \bm{x}_1) & \text{Flow Matching}
\end{align*}
can both generate the same trajectory:
\begin{align*}
\bm{x}_t &= \bm{x}_0 + (\bm{x}_1 - \bm{x}_0)\sqrt{t}
\end{align*}
That being said, DDIM at time $t$ corresponds to flow matching at time $2\sqrt{t}$; thus, flow matching is “slower” than DDIM when $t$ is small. This may be beneficial for flow matching in practice. 
\end{itemize}

\subsection{Activities of Daily Living Dataset}
We construct a real-world dataset with $10$ tasks across Activities of Daily Living. Each task includes $1,000$ sets of RGB-D images, demonstrated robot action trajectories, and labeled ground truth of affordance maps. Thus $10,000$ demonstrations have been collected in total. The data has been manually collected by moving robot end-effectors with kinesthetic teaching. The novelty of our dataset includes: (i) It contains the same scenarios with multiple objects, multi-task affordances, and demonstrated robot trajectories. (ii) All tasks are related to Activities of Daily Living that involve (simulated) human data. 

We label the affordance heatmaps with 2D Gaussian blobs centered on the object pixels of the demonstrated action. The affordance maps model the locations of all relevant object areas that physically interact with robots, given each task. For example, the feeding task requires affordance heatmaps centered on the fork handle, food, and the human mouth.

Objects are randomly placed within the range of the table (1.5 meters $\times$ 1 meter). The human (manikin) is placed randomly around the table in the camera view. We have around 30 different objects. Our tasks include prompt primitives: `sweep the trash', `pass the water to the human', `hang the towel', `put on the hat', `cover the food', `wipe the nose', `wipe the forearm', `feed the human', `comb the hair', and `brush the teeth'. We can also optionally add a pretrained LLM layer (\eg, GPT) in the very front for zero-shot text classification, allowing for linking other language instructions to one of the ten prompt primitives. For example, an ambiguous instruction `I am hungry' can be linked to the prompt primitive `feed the human'. The camera is positioned with some variations but generally oriented towards the table and objects. Please refer to the supplementary materials for more videos of each task.

\section{Experiments}
In this section, we systematically evaluate the performance of the proposed prompt tuning and flow matching methods.

In Section~\ref{sec:PromptEval}, we first benchmark our proposed prompt-tuning structures against several commonly used parameter-efficient fine-tuning approaches to demonstrate the superior performance of prompt tuning. We then compare our method with state-of-the-art VLM-based affordance learning approaches to highlight its efficiency. Additionally, we perform ablation studies to analyze the impact of different design choices on performance. In Section~\ref{sec:FlowEval} and~\ref{sec:RealEval}, we compare our flow-matching policy with state-of-the-art generative learning, imitation learning, and VLA models for robot manipulation through real-world experiments, evaluating performance in terms of generation quality, stability, inference time, and training resource requirements. In Section~\ref{sec:DiffusionEval}, we systematically investigate the performance of flow matching compared to diffusion policy, across simulation and real-world benchmarks.

\subsection{Affordance Evaluation with Prompt Tuning}
\label{sec:PromptEval}

\subsubsection{Baseline Studies}
We benchmark our proposed prompt tuning structures against several commonly used parameter-efficient finetuning protocols and SoA VLM-based affordance learning methods:
\begin{itemize}
[
    \setlength{\IEEElabelindent}{\dimexpr-\labelwidth-\labelsep}
    \setlength{\itemindent}{\dimexpr\labelwidth+\labelsep}
    \setlength{\listparindent}{\parindent}
]
\item Full fine-tuning: fully update the text and vision transformer layers and the decoder.
\item Adapter-based methods: insert MLP layers with residual connections between pretrained frozen transformer layers of vision and language, as customary in the literature~\citep{liu2023tail, sharma2023lossless}.
\item Side-network methods: train a language-based network on the side, append pretrained vision features and sidetuned text features before being fed into the decoder, as customary in the literature~\citep{ganz2024question}. This also shares similarities with parallel adaptors. Our deep prompt tuning method differs in the sense that it inserts text layers into every vision layer, while parallel network methods introduce additional trainable modules that run in parallel with the main transformer layers instead of modifying existing layers.
\item Decoder-based methods: adopt the pretrained backbone as a feature extractor with fixed weights during tuning, and only train the decoder, as customary in the literature~\citep{he2016deep}. 
\item Cross-attention methods: use cross-attention fusing text and vision instead of simple prepending. An example of cross-attention fusing vision and language can be found in the literature~\citep{jiang2022vima}.
\item Mixed-design methods: mixes the favorable adaptor designs. We compare against the state-of-the-art MAM adaptor~\citep{he2021towards}, which is a Mix-And-Match adapter designed based on the practical findings on NLP.
\item VLM-based affordance learning: We also compare against two state-of-the-art affordance learning methods based on fine-tuning large VLMs: AffordanceLLM~\citep{qian2024affordancellm} and Robopoint~\citep{yuan2024robopoint}. 
\end{itemize}

For a fair comparison, all the baselines here use self-supervised pretrained MAE weights on ImageNet-21k dataset for the vision transformer model. We randomly split our Real-world Activities of Daily Living (ADLs) dataset with $80$\%-$20$\% percentage of training and testing. The results reported here are obtained after $1,000$ epochs of training.

\subsubsection{Main Results}

\begin{table}[t]
\resizebox{1.0\columnwidth}{!}{
\begin{tabular}{llrrr}
\toprule[1.5pt]
 & \textbf{Methods} & \textbf{Learnable}     & \textbf{Affordance}  & \textbf{Heatmap}  \\
  && \textbf{Params} $\downarrow$   & \textbf{Heatmaps} ($\times 10^{-3}$) $\downarrow$ & \textbf{Centers} (pixel) $\downarrow$    \\ 
\midrule
Baselines   & Full                              & 153.8M   & \pmb{$0.76$}            & \pmb{$1.15 $}                  \\
            & Decoder                           & 3.9M     & 1.51                    & 13.48                          \\       
            & Adapter                           & 19.2M    & 1.17                    & 6.22                           \\       
            & Cross-attention                   & 43.5M    & 1.26                    & 8.89                            \\
            & side-network                      & 42.7M    & 1.35                    & 9.20                            \\ 
            & mixed-design                      & 108M     & 0.85                    & 2.96                            \\ 
            & AffordanceLLM                     & 7B       & \pmb{$0.72$}            & \pmb{$1.01$}                   \\
            & Robopoint                         & 12B      & -                       & 3.18                            \\
\midrule
Ours        & PT-shallow                        & 8.0M     & $1.42$                  & 12.04                           \\
            & PT-deep (self-supervised weights) & 42.1M    & \pmb{$0.80$}            & \pmb{$2.93$}                     \\ 
\midrule
Ablations   & PT-deep (supervised weights)      & 42.1M    & 1.48                    & 10.13                           \\
            & PT-deep (image output)            & 42.1M    & 1.56                    & 13.27
\\\toprule[1.5pt]
\end{tabular}}
\caption{Results of prompt tuning baseline and ablation studies. We report the number of learnable parameters, the heatmap estimation error (the fourth column) and the heatmap center error (the fifth column). Our method outshines other baselines except for the full finetuning.}
\label{tab:ptbaseline}
\end{table}

\begin{figure}[!t]
\centering
\includegraphics[width=0.98\columnwidth]{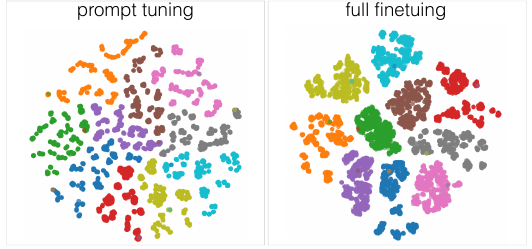}
\caption{t-SNE visualizations of the embeddings before the decoder. The points of the same color denote the tasks with same language prompts, which are embedded together. The prompt tuning method could produce instruction-relevant features without updating vision backbone parameters.}
\label{fig:tsne}
\end{figure}

We use two metrics to evaluate our results: (i) L2 error of affordance heatmap estimation, and (ii) L2 distance between the predicted and ground truth of heatmap centers. We fit Gaussian Mixture Models on predicted heatmaps to determine the inferred heatmap centers. The heatmap error is averaged on each map, and the center error is averaged on per center point. Three observations could be made: 
\begin{itemize}
[
    \setlength{\IEEElabelindent}{\dimexpr-\labelwidth-\labelsep}
    \setlength{\itemindent}{\dimexpr\labelwidth+\labelsep}
    \setlength{\listparindent}{\parindent}
]
\item \textbf{General analysis}: Table~\ref{tab:ptbaseline} presents the results of prompt tuning on our ADLs testing dataset for affordance learning, comparing against baselines. The deep structure of prompt tuning outperforms other parameter-efficient baselines. We can also see that deep prompt tuning achieves better performance than RoboPoint but falls short of the performance achieved by AffordanceLLM. We
would like to mention that AffordanceLLM and Robopoint are not parameter-efficient models (the focus of our research), as they involve finetuning large models, including LLaVA and Vicuna. Full finetuning such a large model may not function optimally if only a small dataset is available.
\item \textbf{Prompt tuning against full finetuning}: Full finetuning slightly outperforms deep prompt tuning in terms of heatmap estimation error and heatmap center error. However, the distinction of heatmap center errors ($1.78$ pixels) remains subtle, given the full image size of $224 \times 224$. This outcome is favorable as it indicates that most heatmap errors are caused by the tails of the Gaussian distribution, instead of the center area where the robot actions are actually applied. We will further ablate the impact of dataset size on these two methods.
\item \textbf{Generalizability}: We also observe that the trained model could be generalized to new objects. For example, the training dataset only includes a manikin. We found out that it generates well on our testing data with real humans. Affordances on objects with similar shapes (\eg, forks and spoons) could also be transferred. Note that as the proposed tuning method is parameter-efficient, it is envisaged that the method could be readily transferred to different tasks with a small amount of task-specific data. Note that the testing experiments involved only the authors as participants and were therefore exempt from the requirement for ethics approval.
\item \textbf{What do prompts learn?} We show a t-SNE~\citep{van2008visualizing} visualization of the embeddings after the last vision transformer layer (before the decoder) in Fig.~\ref{fig:tsne}. We can see that the points of the same color (\eg, tasks with the same language prompts) are embedded together, implying that the representations recover the underlying manifold structure of discriminative task information.
\item \textbf{Prompt tuning or adapters?} As pointed by the research of Visual Prompt Tuning~\citep{jia2022visual}, in contrast to comparable studies in NLP, prompt tuning outperforms full fine-tuning and adapter-based methods in the visual domain. The MAM adaptor~\citep{he2021towards} mixes the favorable adaptor designs based on the practical findings on NLP and achieves state-of-the-art results, but does not function optimally in the image-text domain. 
\end{itemize}

\subsubsection{Ablations} 
We further ablate model design choices:

Pretrained Weights: We evaluate using MAE self-supervised pretrained weights and supervised pretrained weights trained on ImageNet-21k dataset for the vision transformer model. The results in Table~\ref{tab:ptbaseline} show that self-supervised pretrained weights perform better. We are aware of other more complicated variants of vision transformers, for example, CLIP vision encoder and its pretrained weights. As our goal is to integrate textual representations into any vision encoder while keeping it frozen, we have chosen the most basic ViT-B-16 transformer backbone and commonly used pretrained weights and achieved competitive results.

Decoder Input: We apply the decoder on the global output and image-corresponding output after the vision transformer respectively and report results in Table~\ref{tab:ptbaseline}.

\begin{figure}[!t]
\centering
\includegraphics[width=0.98\columnwidth]{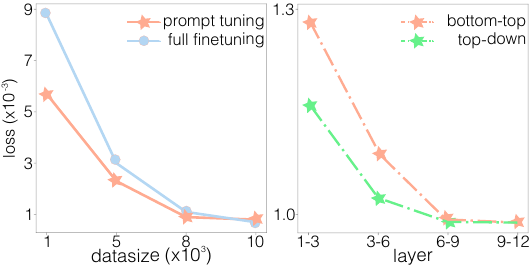}
\caption{Ablation studies of prompt tuning. We investigate the effect of various design choices on affordance learning performance, including pretrained weights, decoder input, dataset size and prompt location.}
\label{fig:ptablation}
\end{figure}

Dataset Size: We use various amounts of data for training. Fig.~\ref{fig:ptablation}-left shows that prompt tuning has better adaptability than full finetuning when downstream data is scarce. 

Prompt Location: We have seen different conclusions from prior works about whether the vision-language fusion should be integrated at early or late transformer layers. We conduct experiments to insert prompts at various layers. From Fig.~\ref{fig:ptablation}-right, we can see that inserting prompts to early layers (for example, layer 1-3 from bottom to top) achieves higher loss than inserting to late layers (for example, layer 1-3 from top to bottom). Thus in our case, prompts have greater significance at the late transformer layers. These results are also supported by the nature of the vision transformer hierarchy: lower layers mainly capture low-level fundamental visual details, while higher layers focus on high-level concepts that might be vital for downstream tasks.

In conclusion, we observe no single method that outperforms all the rest. For scenarios where a small number of parameters or datasize is available, we reckon that prompt tuning remains the preferred approach.

\subsection{Flow Matching Policy Evaluation}
\label{sec:FlowEval}

\subsubsection{Baseline Studies}
We compare our flow matching policy against: (i) Diffusion Policy~\citep{chi2023diffusion} with DDPM and DDIM, (ii) Transformer-based behavior cloning, as customary in RVT~\citep{goyal2023rvt}, RT-X~\citep{padalkar2023open}, and (iii) Score-based generative models~\citep{song2020score} in Section \ref{sec:DiffusionEval}.

Note that we are aware of other competitive robot behavior cloning methods, including energy-based IBC~\citep{florence2022implicit}, GAIL~\citep{ho2016generative}, etc. Since extensive studies have been conducted and showed better performance of the diffusion policy against these methods, we choose the representative transformer and diffusion baselines for evaluation. Table~\ref{tab:hyperparameters} shows the hyperparameters used in flow matching and diffusion policy.

We first train and test flow matching and baselines on our collected Real-world Activities of Daily Living (ADLs) dataset in a supervised manner with Mean Square Error Loss. In the following sections, we will also evaluate on other benchmarks (\eg, Push-T, Franka Kitchen, and Robomimic).  
\begin{itemize}
[
    \setlength{\IEEElabelindent}{\dimexpr-\labelwidth-\labelsep}
    \setlength{\itemindent}{\dimexpr\labelwidth+\labelsep}
    \setlength{\listparindent}{\parindent}
]
\item \textbf{ADLs benchmark}: For 2D data, the training uses a RGB image with visual affordances as input, and the output is a trajectory in 2D pixel space. For the counterparts in 3D, the training takes the concatenation of RGB-D images with visual affordances as input, and outputs a trajectory in 3D Cartesian space. Since the ADLs task is not closed-loop, as shown in Table~\ref{tab:taskparameters}, we feed the image with all task-related object affordances to the flow matching policy, which directly outputs the entire trajectory. We randomly split the dataset with $80$\%-$20$\% percentage of training and testing. The results are reported on the testing dataset, with each experiment conducted one time, after $1,000$ epochs of training. For each baseline, all 10 tasks in our dataset are trained in a single policy in a supervised manner. To ensure fair comparisons, the corresponding hyperparameters across the flow matching policy and baseline methods are selected to remain consistent. Table~\ref{tab:taskparameters} shows the task summary.

\end{itemize}

\begin{table}[t]
\resizebox{0.99\columnwidth}{!}{
\begin{tabular}{llrrr}
\toprule[1.5pt]
 & \textbf{Methods} & \textbf{2D Trajectory}     & \textbf{3D Trajectory}  & \textbf{Inference}  \\
  && \textbf{Prediction} (pixel) $\downarrow$  & \textbf{Prediction} (cm) $\downarrow$ & \textbf{Times} (ms) $\downarrow$ 
\\\midrule
Ours        & Flow Matching (Transformer, 16-step)  & 1.061           & 1.239         & 140.71              \\
            & Flow Matching (CNN, 16-step)     & \pmb{$0.840$}        & \pmb{$1.009$} & 98.981                   
\\\midrule 
Ablations   & Flow Matching (CNN, 1-step)      & 0.888                & 1.031    & 13.228                   \\ 
            & Flow Matching (CNN, 4-step)      & 0.846                & 1.014    & 41.494                   \\
            & Flow Matching (CNN, 8-step)      & 0.842                & 1.013    & 78.222                   
\\\midrule 
Baselines   & DDPM (1-step)                    & 2.851                & 2.479    & 13.791                   \\    
            & DDPM (4-step)                    & 0.890                & 2.411    & 45.736                   \\   
            & DDPM (8-step)                    & 0.884                & 2.403    & 80.200                   \\    
            & DDPM (16-step)                   & 0.882                & 2.398    & 99.197                   \\  
            & DDIM (1-step)                    & 4.328                & 10.78    & 13.757                   \\    
            & DDIM (4-step)                    & 0.876                & 2.299    & 42.672                   \\   
            & DDIM (8-step)                    & 0.875                & 2.272    & 80.940                   \\    
            & DDIM (16-step)                   & 0.874                & 2.266    & 98.680                   \\   
            & Transformer-based BC             & 2.797                & 4.911    & 7.59                                      
\\\toprule[1.5pt]
\end{tabular}}
\caption{Results of flow matching policy against baselines and ablations. We report the average error of 2D and 3D trajectory estimation and the inference time. Our flow matching method achieves the best trajectory estimation accuracy. We also investigate the effect of various design choices on flow matching performance, including network structures, training, and inference steps.}
\label{tab:fmbaseline}
\end{table}

\subsubsection{Main Results}
Table~\ref{tab:fmbaseline} presents the results of flow matching policy on our ADLs testing dataset for robot trajectory learning, comparing against baselines. We use two metrics for evaluation: (i) the average error of 2D and 3D trajectory estimation, and (ii) the average inference time, performed with RTX 4090 GPU. The trajectory error is averaged on each point of the trajectory.

\begin{figure}[!t]
\centering
\includegraphics[width=0.98\columnwidth]{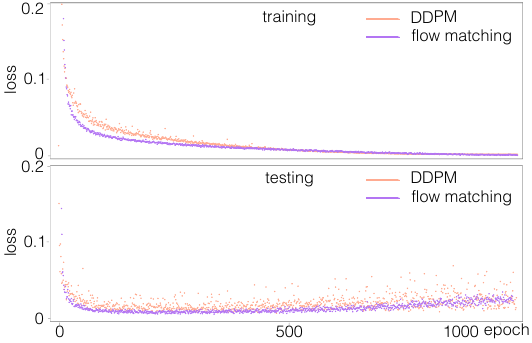}
\caption{Training and testing loss throughout the training process. Flow matching exhibits greater stability on both training and evaluation than diffusion policy.}
\label{fig:loss}
\end{figure}

Four observations could be made from this result: 
\begin{itemize}
[
    \setlength{\IEEElabelindent}{\dimexpr-\labelwidth-\labelsep}
    \setlength{\itemindent}{\dimexpr\labelwidth+\labelsep}
    \setlength{\listparindent}{\parindent}
]
\item \textbf{Generation quality:} Flow matching (CNN-based, 16 steps) outperforms diffusion policy and Transformer baselines in terms of 2D and 3D trajectory prediction accuracy. The suboptimal performance of Transformer behavior cloning is expected as it is hindered by the nature of multi-modal action distribution, causing the averaging out across non-convex spaces.

\item \textbf{Stability:} Fig.~\ref{fig:loss} shows an example of training and testing loss of flow matching and diffusion policy with DDPM throughout the training process. We can see \textbf{flow matching exhibits greater stability on both training and evaluation than diffusion policy}.

\item \textbf{Inference time:} We have two observations here: i) Table~\ref{tab:fmbaseline} showcases that flow matching with 16 steps achieves faster inference time compared to diffusion policy with 16 steps. We hypothesize that flow matching with linear pointwise flows generates straighter flows than DDPM and DDIM, and thus causes faster inference. ii) More importantly, Table~\ref{tab:fmbaseline} also showcases that 1-step flow matching (error: $1.031$cm, time: $13.228$ms) has achieved comparable performance as 16-step DDIM (error: $2.266$cm, time: $98.680$ms), but \textbf{noticeably lowered inference time roughly by 85\%}. We hypothesize that this is because diffusion models solve a stochastic differential equation with a series of discrete steps to progressively refine the generated sample. Contrarily, flow matching trains continuously normalize flow models, leading to no significant improvements when increasing inference steps. Thus 2-step flow matching has achieved comparable performance as the 16-step diffusion policy, which considerably reduces the inference time for closed-loop robot manipulation. This is \textbf{in line with the results obtained in the image generation domain}. As pointed out by Stable Diffusion 3~\citep{esser2024scaling}, flow matching outperforms diffusion policy with fewer inference steps, making it particularly advantageous in scenarios that demand fast inference.

\item \textbf{Training resources:} DDPM training and benchmarking demand significant resources for various training and inference steps. DDIM decouples the number of denoising iterations in training and inference, thereby allowing the algorithm to be trained one time with a large training iteration and use fewer iterations for inference to speed up the process. However, flow matching still achieves faster inference than DDIM.
\end{itemize}

\subsubsection{Ablations} 
We further ablate policy design choices.
\begin{figure}[!t]
\centering
\includegraphics[width=1.0\columnwidth]{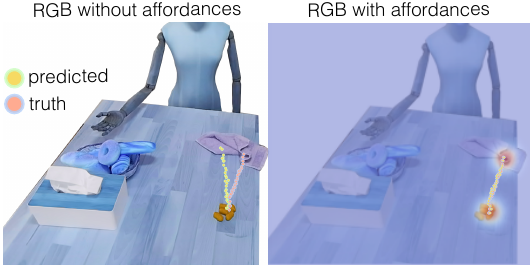}
\caption{Ablations of using RGB images with/without affordances for policy training. Visual affordance guides the flow matching policy to generate a trajectory closely aligned with the truth. The policy without affordances might generate a trajectory detached from the ground truth, but still a reasonable solution. This reinforces the argument that flow matching policy could handle multimodal robot action distributions.}
\label{fig:robotmultimodal}
\end{figure}

Network Structure: As shown in Table~\ref{tab:fmbaseline}, CNN-based flow matching achieves better results than transformer-based architecture. We hypothesize that transformer might need additional hyperparameter tuning.

Trajectory Representation: We empirically test trajectory representation with 8, 16, 32 and 64 waypoints. More waypoints are not necessary, while fewer waypoints are unable to entirely encapsulate the complete long-horizon trajectories. We have found that, in general, the trajectory representation does not wield a significant influence on flow matching performance. The performance reported in the above main results section is achieved by using 32 waypoints.

\subsection{Real-World Robot Evaluation}
\label{sec:RealEval}

We deploy our flow matching policy, DDPM and Transformer policies on real robot manipulation for evaluation. We carry out 50 replications of trials for each baseline. We use a KINOVA Gen3 arm and an Azure Kinect camera for real-world robot experiments. Details can be found in the supplementary video. Besides, we have respectively compared the proposed parameter-efficient affordance learning and flow matching policy against SoA baselines. In this section, we also investigate whether our framework can seamlessly unify affordance learning and action generation. 

\textbf{Affordance-based policy}: From Table~\ref{tab:real}, we can see that flow matching outperforms DDPM and Transformer baselines.

\textbf{End-to-end learning (RGB images without affordances)}: We investigate how the affordance would guide the flow matching policy. We trained flow matching taking the raw RGB images and language tokens as input. This is similar to our proposed method but without the intermediate stage of affordance learning. It also involves some resemblance while not being entirely identical to the method in~\citep{black2024pi_0} in an end-to-end learning manner. Interestingly, comprehensive examinations reinforce the argument that flow matching could handle multimodal action distribution. Fig.~\ref{fig:robotmultimodal} shows one example. From the left figure, we can see that when training a policy without affordances, the predicted trajectory (yellow) of moving the towel toward the trash for sweeping could be detached from the ground truth (red), but still a reasonable solution that allows for a successful robot execution. With affordance guidance, the prediction is closely aligned with the truth (Fig.~\ref{fig:robotmultimodal}-right, Table). We also observe that for applications that demand higher precision in manipulation, like grasping the toothbrush handle, affordances offer greater guidance for shaping the manipulation policy.

\textbf{Off-the-shelf VLM-based trajectory generation}: Our method uses an extra low-level module (flow matching) to learn a robot policy built on the high-level VLM reasoning. We have also conducted preliminary research to test another popular line of research, which uses a VLM to generate robot actions directly, as customary in the literature~\citep{huang2023voxposer, tanneberg2024help}. No constant performance was achieved without carefully designed prompts or finetuning. This result is expected as several recent state-of-the-art research (e.g., HAMSTER~\citep{li2025hamster}, $\pi$0~\citep{black2024pi_0}) have shown that VLA models with robot learning modules outperform VLM-based trajectory generation.

\begin{table}[!t]
\centering
\resizebox{1.0\columnwidth}{!}{
\begin{tabular}{lrrrr}
\toprule[1.5pt]
\textbf{Methods}               & \textbf{Flow Matching}           & \textbf{Diffusion Policy}       & \textbf{Transformer}      & \textbf{Flow Matching} \\
(\textbf{Inference Step})      & \textbf{($16$-step)} $\uparrow$  & \textbf{($16$-step)} $\uparrow$ & \textbf{BC}  $\uparrow$   & \textbf{end-to-end}  $\uparrow$                   
\\ \midrule
Activities of Daily Living     & \pmb{$0.82$}       & 0.76      & 0.44     & 0.74                                            
\\\toprule[1.5pt]
\end{tabular}}
\caption{Real-world robot experimental results.}
\label{tab:real}
\end{table}

\begin{table}[!t]
\centering
\includegraphics[width=0.99\columnwidth]{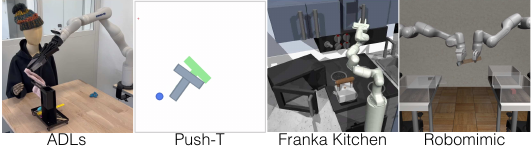}
\vskip 0.1in
\resizebox{1.0\columnwidth}{!}{
\begin{tabular}{lrrrrr}
\toprule[1.5pt]
\textbf{Methods (16-step)}  & \textbf{Push-T}$^a$ $\uparrow$    & \textbf{Push-T}$^b$ $\uparrow$      & \textbf{Franka Kitchen} $\uparrow$ & \textbf{Robomimic} $\uparrow$   
\\\midrule
Flow Matching     & \pmb{$0.9035/0.7519$} & 0.7363/\pmb{$0.6218$}    & \pmb{$0.9960$}/0.7425   & 0.9360/\pmb{$0.7289$}  \\
DDPM              & 0.8840/0.7178         & 0.7360/0.6100            & 0.9840/0.6716           &  0.9359/0.7168          \\
DDIM              & 0.8801/0.6372         & 0.7490/0.6167    & 0.9865/\pmb{$0.7471$}   &  0.9334/0.7073          \\
Score-based       & 0.8950/0.6432         & \pmb{$0.7659$}/0.6186            & 0.9863/0.6892           &  \pmb{$0.9470$}/0.4896
\\\toprule[1.5pt]
\end{tabular}}
   \begin{tablenotes}[para,flushleft]
    \item[a] sampling range: [(50, 450), (50, 450), (200, 300), (200, 300), ($-\pi, \pi$)] \\
    \item[b] sampling range: [(50, 450), (50, 450), (100, 400), (100, 400), ($-\pi, \pi$)] \\
    \end{tablenotes}
\caption{We present the robot evaluation performance in the format of (max performance) / (average of last checkpoint with $10$ trials of replications), with each averaged across $500$ different environment initial conditions. The metric used here is success rate, except for the Push-T task which uses target area coverage. We have used various sampling ranges (end-effector position, T-block position and orientation) for Push-T environment initialization. For Robomimic benchmark, we specifically report results on the Transport task.}
\label{tab:bench}
\end{table}

\begin{table*}[!t]
\centering
\resizebox{1.0\textwidth}{!}{
\begin{tabular}{lrrrrrrrrrrrrrr}
\toprule[1.5pt]
\textbf{H-Param}                      & \textbf{Ta}      & \textbf{Tp}           & \textbf{ObsRes}       & \textbf{F-Net}                            & \textbf{F-Par}
& \textbf{V-Enc}                      & \textbf{V-Par}  & \textbf{Lr}           & \textbf{WDe}       & \textbf{Iters Train (FM)}      & \textbf{Iters Train (DDPM)}  & \textbf{Iters Train (DDIM)}                     & \textbf{Iters Eval}    
\\\midrule
Activities of Daily Living          & 32               & 32                    & 1x224x224               & ConditionalUnet1D & 72                    & ResNet-18             & 11                & 1e-4                                & 1e-6             & N/A  & 1/4/8/16 & 16           & 1/4/8/16              \\
Push-T                                & 8                & 16                    & 1x96x96                 & ConditionalUnet1D & 80                    & ResNet-18             & 11                & 1e-4                                & 1e-6             & N/A  & 1/4/8/16 & 16           & 1/4/8/16              \\
Franka Kitchen                        & 8                & 16                    & 1x60                   & ConditionalUnet1D & 66                    & N/A                   & N/A               & 1e-4                                & 1e-6             & N/A  & 1/4/8/16 & 16           & 1/4/8/16              \\
Robomimic                             & 8                & 16                    & 1x50                   & ConditionalUnet1D & 66                    & N/A                   & N/A               & 1e-4                                & 1e-6             & N/A  & 1/4/8/16 & 16           & 1/4/8/16             
\\\toprule[1.5pt]
\end{tabular}}
\caption{Hyperparameters for flow matching and diffusion policy. Ta: action horizon. Tp: action prediction horizon. ObsRes: environment observation resolution. D-Net: diffusion/flow matching network. D-Par: diffusion/flow matching network number of parameters in millions. V-Enc: vision encoder. V-Par: vision encoder number of parameters in millions. Lr: learning rate. WDe: weight decay. Iters Train: number of training diffusion iterations. Iters Eval: number of inference iterations.}
\label{tab:hyperparameters}
\end{table*}

\begin{table}[!t]
\centering
\resizebox{1.0\columnwidth}{!}{
\begin{tabular}{lrrrrrrr}
\toprule[1.5pt]
\textbf{Tasks}   & \textbf{Rob} & \textbf{Obj} & \textbf{ActD} & \textbf{PH}  & \textbf{Steps} & \textbf{Img}   & \textbf{Closed-loop}     
                 \\ \midrule
Activities of Daily Living    & 1            & $\approx 30$  & 3            & 8,000          & N/A            & \cmark         & \xmark        \\
Push-T           & 1            & 1            & 2             & 200          & 300            & \cmark         & \cmark          \\
Franka Kitchen   & 1            & 7            & 9             & 566          & 280            & \xmark         & \cmark          \\
Robomimic        & 2            & 3            & 20            & 200          & 700            & \xmark         & \cmark     
\\\toprule[1.5pt]
\end{tabular}}
\caption{Tasks Summary. Rob: number of robots. Obj: number of objects. ActD: action dimension. PH: proficient-human demonstration. Steps: max number of rollout steps. Franka Kitchen and Robomimic involve 6D robot and gripper actions in the joint space. ADLs and Push-T focus on robot end-effector trajectories. For clarity, we further explain that for our ADLs tasks, as no closed-loop motion is considered here, we assume that the gripper closes when the first waypoint has arrived, and the low-level actions are executed between waypoints using a standard proportional-derivative (PD) controller.}
\label{tab:taskparameters}
\end{table}

\subsection{Comparisons between flow matching and diffusion policy}
\label{sec:DiffusionEval}

To further investigate the performance of flow matching compared to diffusion policy, we benchmark the proposed methods on three more datasets which include closed-loop 6D robot actions and gripper actions: (i) Push-T~\citep{florence2022implicit}, (ii) Franka Kitchen~\citep{gupta2019relay}, and (iii) Robomimic~\citep{robomimic2021}. 
\begin{itemize}
[
    \setlength{\IEEElabelindent}{\dimexpr-\labelwidth-\labelsep}
    \setlength{\itemindent}{\dimexpr\labelwidth+\labelsep}
    \setlength{\listparindent}{\parindent}
]
\item \textit{Push-T} requires pushing a T-shaped block to a fixed target with a circular end-effector. Push-T takes RGB images with proprioception of end-effector location as inputs, and outputs end-effector actions in a closed-loop manner. The dataset includes $200$ demonstrations. 
\item \textit{Franka Kitchen} contains 7 objects for interaction and comes with a human demonstration dataset of $566$ demonstrations, each completing 4 tasks in arbitrary order. The goal is to execute as many demonstrated tasks as possible, regardless of order. The training takes state-based inputs and outputs closed-loop robot joint actions and gripper actions.
\item \textit{Robomimic} consists of 5 tasks with a proficient human teleoperated demonstration dataset. We specifically focus on the transport task which includes $200$ demonstrations. The policy takes state-based inputs and outputs closed-loop robot joint actions and gripper actions.
\end{itemize}

For each benchmark, the evaluation has been carried out across $500$ different environment initial conditions, using the last checkpoint of each policy with $10$ trials of replications. Thus, $5,000$ trials have been carried out in total per policy and benchmark. Variation is added on random initial conditions for the robot and object states. We respectively report the best and average performance in the $10$ trials of replications of the last checkpoint. All state-based tasks are trained for $4,500$ epochs, and image-based tasks for $3,000$ epochs. 

\citep{chi2023diffusion} has reported that setting prediction and action horizons greater than 1 helps the policy generate more consistent actions and compensate for idle portions of demonstrations; however, excessively long horizons can degrade performance due to slower reaction times. To ensure fair comparisons, we have selected the same corresponding hyperparameters across the flow matching policy and baseline methods to remain consistent. Table~\ref{tab:hyperparameters} shows the hyperparameters we have used in flow matching and diffusion policy. Table~\ref{tab:taskparameters} shows the task summary.

Similar conclusions could be achieved from Table~\ref{tab:bench} and Fig.~\ref{fig:step}, as in the Main Results section: 
\begin{itemize}
[
    \setlength{\IEEElabelindent}{\dimexpr-\labelwidth-\labelsep}
    \setlength{\itemindent}{\dimexpr\labelwidth+\labelsep}
    \setlength{\listparindent}{\parindent}
]
\item \textbf{Generation quality:} The performances of flow matching and DDPM, DDIM and the score-based model are comparable, where flow matching performs marginally better in most cases. We observe that score-based models can achieve higher peak performance, but their average performance tends to be less consistent.

\item \textbf{Inference time:} Fig.~\ref{fig:step} shows how the number of inference steps affects the performance of flow matching and diffusion policy. We can observe that diffusion policy showcases better performances when applying more inference iterations with a trade-off of longer inference time, as it requires a series of discrete steps to progressively refine the generated sample (solid line). Score-based models with black-box numerical ODE samplers, as in the original research paper~\citep{song2020score}, need adaptive steps to achieve optimal results (around $100-300$). Contrarily, flow matching has not shown significant improvements when increasing inference steps. From 8 steps onward, the performance of both flow matching and the diffusion policy increases only slightly with additional steps. However, the inference time (dotted line, we use frequency here) increases proportionally with the number of inference steps (dotted line). Therefore, flow matching considerably reduces the inference time for closed-loop robot manipulation. In the Push-T benchmark in Fig.~\ref{fig:step}, 2-step flow matching (coverage: $0.8803$, time: $13.098$ms) has achieved comparable performance as 16-step diffusion policy with DDIM (coverage: $0.8801$, time: $98.268$ms), \textbf{but noticeably lower inference time roughly by 86\%}. 
\end{itemize}

Based on the above experimental observations, we conclude that flow matching formulates the robot policy generation problem as learning a deterministic or near‐deterministic vector field (via an ordinary differential equation) that maps noise (or a simple prior) to target action distributions—this ODE-based framing avoids iterative stochastic denoising chains (as in diffusion policy) and yields fast, single- or few-step inference, which is crucial in closed-loop robotic settings. Besides, in manipulation tasks, the target action distribution is often multimodal (e.g., multiple possible grasp poses, multiple contact sequences) but still has strong structure (e.g., physics constraints, smooth trajectories through state‐action space). Flow matching’s vector‐field representation permits modeling such multimodality while preserving smooth interpolation and strong inductive bias towards consistent flows—thus potentially better capturing the structured trajectory manifold than, say, a GAN which may collapse modes, or a diffusion model which requires many steps. Moreover, because robotic manipulation is extremely sensitive to inference latency, stability, and feedback responsiveness, flow matching’s fewer-step generation is advantageous. The explicit mapping of a vector field lends itself more naturally to feedback correction and incorporation of geometric/physical structure (e.g., SE(3) equivariance) than purely stochastic models. Overall, flow matching combines fast inference and structured multimodal representation, making it a compelling choice for manipulation domains.

During our experiments, we observed that most failures were caused by out-of-distribution factors. In real-world experiments, variations such as background changes or significant camera view shifts often led to failures of the 3D flow-matching policy. Increasing both the dataset size and diversity could help mitigate this issue. Unlike the simulation benchmarks, we performed open-loop manipulation in the real world, meaning that a failed first grasp attempt was counted as a failure; closed-loop manipulation could potentially address this limitation. We also found that tasks requiring higher manipulation precision, such as grasping a toothbrush handle, exhibited a higher probability of failure. In simulation benchmarks, out-of-distribution failures were also observed. For example, in the Push-T benchmark, only successful trajectories were included during training, so the policy struggled with edge cases, such as a T-block stuck in a corner. Extending the imitation policy to an online or offline reinforcement learning framework may help address these limitations.

\section{Limitations} 
In this section, we provide a detailed analysis of the advantages and limitations of our proposed affordance-based flow matching approach.

\textbf{Inference time:} The generation quality of flow matching and diffusion policy for robot manipulation are generally comparable. Although we see only a marginal improvement in flow matching in most cases, we would like to highlight that the focus of this work is not to outperform state-of-the-art general robot manipulation research. Instead, we have systematically studied the flow matching framework, which provides an alternative to diffusion policies for robot manipulation. We can not overlook the additional advantages of flow matching, including stable training, easy implementation, and most importantly, significantly better performance and faster inference with fewer inference steps than diffusion policy, suggesting forsaking the stochastic construction of diffusion policy in favor of learning the probability path more directly as in flow matching.

\begin{figure}[!t]
\centering
\includegraphics[width=0.98\columnwidth]{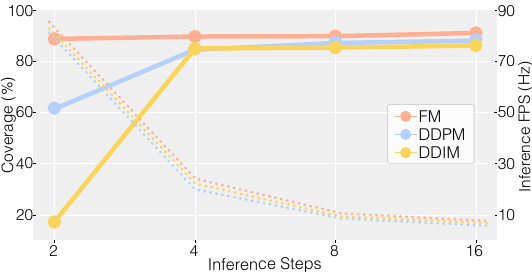}
\caption{Generation quality (solid line) and inference time (dotted line, we use frequency here) comparison of flow matching and diffusion policy for varying values of inference steps.}
\label{fig:step}
\end{figure}

We have evaluated how varying inference steps affect the performance of flow matching and diffusion policy. The original diffusion policy research~\citep{chi2023diffusion} uses a large $100$ training and inference steps with DDPM in their experiments. Based on our evaluation in Fig.~\ref{fig:step} and results from other research~\citep{chisari2024learning}, we can observe that beyond $8$ steps, further increasing steps have only a marginal impact on the performance of diffusion policy, but with a trade-off of significantly longer inference time. We are aware of recent research on one-step diffusion policy with distillation~\citep{prasad2024consistency} and shortcut models~\citep{frans2024one}. We primarily focus on conducting a comparative analysis of the fundamental architectures underlying flow matching and vanilla diffusion policy with DDPM, DDIM and the score-based model.

\textbf{Deterministic and stochastic tasks:} Flow Matching is inherently more deterministic than diffusion-based approaches, which can have implications in highly stochastic tasks—i.e., tasks where the same actions can lead to different outcomes due to environmental randomness or uncertainty, such as manipulating deformable objects, pushing objects on uneven surfaces, or navigating in dynamic environments. The impact of flow matching’s determinism also depends on the training dataset: with a diverse dataset covering a wide range of outcomes, flow matching can learn to predict the most likely trajectories effectively, whereas in limited or biased datasets, flow matching may fail to capture rare but plausible outcomes. 

\textbf{Trajectory smoothness:} Although our flow matching framework effectively learns Cartesian-space or joint-space trajectory representations, it does not explicitly enforce second-order smoothness or torque feasibility during training or inference. In practice, the trajectories are executed through standard robot APIs (e.g., Kinova), which apply spline-based interpolation and internal velocity/acceleration limits to ensure dynamically feasible and smooth motion. Nevertheless, incorporating explicit smoothness or dynamic-feasibility constraints—such as velocity or acceleration regularization terms, or post-processing filters—could further improve motion consistency and physical plausibility. However, as noted in recent works~\citep{romer2024diffusion, zeng2025optimization}, these additions typically increase computational complexity, highlighting a trade-off between control fidelity and real-time efficiency. Exploring this direction is a valuable avenue for future research.

\textbf{Transportation cost:} While our method successfully leverages basic flow matching to generate robust and feasible affordance-based trajectories with a high success rate, we acknowledge that it does not explicitly pursue the minimal geometric transportation cost. More advanced flow techniques, such as Rectified Flow Matching (RFM)~\citep{liu2022rectified} or methods based on Optimal Transport (OT)~\citep{villani2021topics}, are mathematically designed to produce straight-line mappings that minimize this cost. However, implementing these methods introduces significant training and preprocessing overhead, often necessitating complex iterative or global optimization schemes. Given that our primary goal is to explore the fundamental structural efficacy and computational tractability and use affordances to shape the target distribution, we adopt the more efficient basic flow matching objective. We are deferring the complex, multi-objective optimization that simultaneously constrains geometric factors like path length and crucial practical constraints to future work.

\textbf{Scalability:} The current scope of our work presents several avenues for future generalization. The ADLs dataset was constructed in a controlled setting. We acknowledge this limited variation necessitates future work in demonstrating cross-domain generalization (e.g., varying lighting or camera poses), as acquiring and annotating the vast, diverse datasets required for open-world robustness is often infeasible for typical research settings. Furthermore, while our parameter-efficient prompt tuning leverages the semantic robustness of a pretrained foundation model to learn manipulation affordances, incorporating the spatial and semantic interactions between elements in multi-task scenarios, its ultimate performance against extreme linguistic noise is fundamentally data-driven. We plan to address these challenges in future work by expanding dataset diversity and evaluation rigor. Besides, we are exploring future extensions that incorporate additional modalities (e.g., haptic or force-torque sensing) or dynamics models to provide a richer, dynamics-aware affordance representation.


\section{Conclusion} 
We have formulated a prompt tuning method for affordance map learning and flow matching policy for robot manipulation. The core idea of prompt tuning is to maximally exploit the pretrained foundation model, and rapidly excavate the relevance of foundation and downstream affordance learning tasks. We have proposed a flow matching policy constructing paths that allow faster inference, and improved generation amongst robot behavior cloning methods. We qualitatively and quantitatively experiment on multiple robot manipulation benchmarks to prove that flow matching produces better trade-offs between computational cost and sample quality compared to prior competing diffusion-based methods.

\bibliographystyle{plainnat}
\bibliography{references}

\end{document}